\documentclass[sigconf,natbib=true]{acmart}
\AtBeginDocument{%
  \providecommand\BibTeX{{%
    \normalfont B\kern-0.5em{\scshape i\kern-0.25em b}\kern-0.8em\TeX}}}

\usepackage{graphicx}
\usepackage{mathtools}
\usepackage{wrapfig}
\usepackage{amsfonts}
\usepackage{multirow}
\usepackage{colortbl}
\usepackage{amsmath}
\usepackage{caption}
\usepackage[ruled,vlined,linesnumbered]{algorithm2e}

\renewcommand\footnotetextcopyrightpermission[1]{} 

\begin{document}

\title{FiSH: Fair Spatial Hot Spots}

\author{
  Deepak P}
\affiliation{
  \institution{Queen's University Belfast} 
  \country{United Kingdom}}
\email{
  deepaksp@acm.org
}
\author{Sowmya S Sundaram}
\affiliation{\institution{L3S Research Center} \country{Germany}}
\email{sundaram@l3s.de}

\begin{abstract}
Pervasiveness of tracking devices and enhanced availability of spatially located data has deepened interest in using them for various policy interventions, through computational data analysis tasks such as spatial hot spot detection. In this paper, we consider, for the first time to our best knowledge, fairness in detecting spatial hot spots. We motivate the need for ensuring fairness through statistical parity over the collective population covered across chosen hot spots. We then characterize the task of identifying a diverse set of solutions in the noteworthiness-fairness trade-off spectrum, to empower the user to choose a trade-off justified by the policy domain. Being a novel task formulation, we also develop a suite of evaluation metrics for fair hot spots, motivated by the need to evaluate pertinent aspects of the task. We illustrate the computational infeasibility of identifying fair hot spots using naive and/or direct approaches and devise a method, codenamed {\it FiSH}, for efficiently identifying high-quality, fair and diverse sets of spatial hot spots. FiSH traverses the tree-structured search space using heuristics that guide it towards identifying effective and fair sets of spatial hot spots. Through an extensive empirical analysis over a real-world dataset from the domain of human development, we illustrate that FiSH generates high-quality solutions at fast response times. 

\end{abstract}

\maketitle
\pagestyle{plain}

\section{Introduction}

With sensing and tracking devices such as mobile phones and IoT becoming pervasive in this web-driven era, there is an abundance of spatial data across real-world settings. Within such spatial datasets, it is often of interest to identify geographically localized groups of entities that are of sufficient size and express a distinctive character so strongly that it is unlikely to have occurred by chance. To illustrate an example from our times, COVID-19 contact tracing apps accumulate large amounts of spatial data of people, of which some are known to have a COVID-19 infection. It would be of interest to automatically identify localized regions of high COVID-19 incidence, referred to as {\it hot spots} in contemporary reporting\footnote{https://www.nbcnews.com/news/us-news/map-track-summer-2020-coronavirus-hotspots-united-states-n1231332}, so that the information could be channelized to health experts to identify causal reasons, or to public policy experts to develop a mitigation strategy for those regions. 

\begin{figure}
  \centering
  \includegraphics[width=0.4\textwidth]{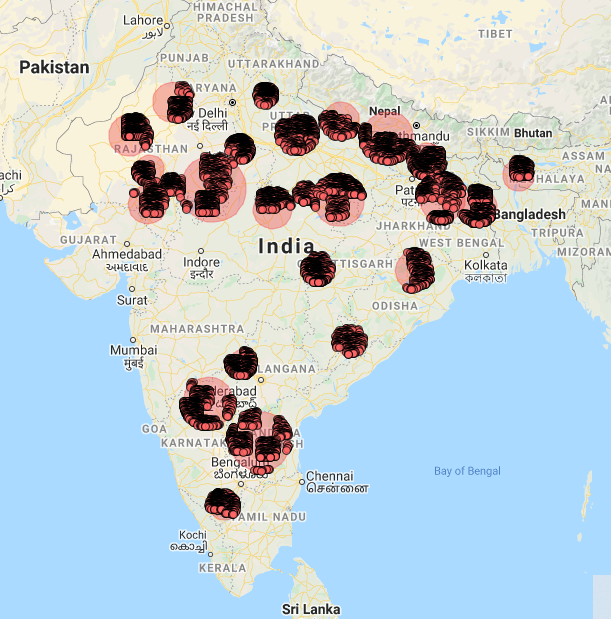}
  \caption{An illustration of hot spots of Low Educational Achievement in India}
  \label{fig:hotspot}
\end{figure}

While COVID-19 hot spots are characterized by {\it high disease incidence rates}, other web and new age data scenarios may call for different formulations of hot spot character, viz., {\it high crime rates} (law enforcement), {\it intense poverty} (development studies), {\it high mobile data usage} (mobile network optimization) and so on. For example, Figure~\ref{fig:hotspot} illustrates hot spots of educational underachievement in India as identified from a human development dataset. In each case, identifying a set of hot spots would be of use so they could be subjected to an appropriate policy action. The unsupervised learning task of detecting spatial hot spots was pioneered by the spatial scan statistic (SSS)~\cite{kulldorff1997spatial}. The spatial scan statistic and its variants within the SaTScan\footnote{https://www.satscan.org/} toolkit have remained extremely popular for detecting spatial hot spots over the past two decades. While health and communicable diseases form the most popular application area of SSS (e.g.,~\cite{pinchoff2015spatial}), they have been used within domains as diverse as archaeology~\cite{wilczek2015unsupervised} and urban planning~\cite{helbich2012beyond}. 

\subsection{Fairness in Hot Spots}\label{sec:fairhotspots}

In scenarios where spatial hot spots are to be used to inform government and public sector action, especially in sensitive policy domains (e.g., law enforcement~\cite{mohler2018penalized}, development), it is often important to ensure that the collective population subject to the policy action is diverse in terms of protected attributes such as ethnicity, caste, religion, nationality or language, among others. 

Consider hot spot detection on a crime database to inform policy action that could include sanctioning higher levels of police patrols for those regions. This would likely lead to higher levels of {\it stop and frisk} checks in the identified hot spots, and would translate to heightened inconvenience to the population in the region. Against this backdrop, consider a sensitive attribute such as ethnicity. If the distribution of those who land up in crime hot spots is skewed towards particular ethnicities, say minorities as often happens~\cite{meehan2002race}, it directly entails that they are subject to much more inconvenience than others. These,  and  analogous  scenarios  in  various  other  sectors,  provide  a  normatively compelling case to ensure that the inconvenience load stemming from crime hot spot detection (and other downstream processing) be proportionally distributed across ethnicities. The  same  kind of  reasoning  holds  for  groups defined over other  sensitive  attributes such as religion and nationality. It is also notable that ethnically skewed hot spot detection and patrolling would exacerbate the bias in future data. Minor crimes are recorded in the data only when committed as well as observed. Thus, majority and minority areas with similar real crime prevalance, alongside minority-oriented patrolling, would lead to data that records higher crime prevalance in the latter. Second, even in cases where the intended policy action is positive (e.g., setting up job support centres for unemployment hot spots), the policy being perceived as aligned to particular ethnicities could risk social solidarity and open avenues for populist backlash~\cite{greven2016rise}, which could ultimately jeopardize the policy action itself.

While considerations as above are most felt in policy domains such as policing and human development, these find expression in hot spot based prioritization in provisioning any common good. Ensuring fair distribution of the impact of any policy action, across sensitive attributes such as ethnicities, is aligned with the theory of {\it luck egalitarianism}~\cite{knight2009luck}, one that suggests distributive shares (of inconvenience or benefits) be not influenced by arbitrary factors, especially those of ‘brute luck’ that manifest as membership in sensitive attribute groups such as ethnicity, religion and gender (since individuals do not choose those memberships are are often just {\it born into} one). Such notions have been interpreted as a need for orthogonality between groups in the output and groups defined on sensitive attributes, and has been embedded into machine learning algorithms through the formulation of {\it statistical parity} (e.g.,~\cite{DBLP:conf/edbt/Abraham0S20}). 

In summary, there is an compelling case, as in the case of other machine learning tasks, for {\it hot spot detection to be tailored in a way that the population covered across the chosen hot spots be diverse along protected demographic groups} such as ethnicity, gender religion, caste and similar. 


\subsection{Our Contributions}

We now outline our contributions in this paper. {\it First}, we characterize the novel task of detection of fair spatial hot spots, for the first time. In particular, we outline a task formulation for enumerating a diverse sample of trade-off points in the noteworthiness-fairness spectrum, to suit diverse scenarios that require different trade-off points between noteworthiness and fairness. We note that straightforward solutions for the task would be computationally infeasible for even moderate dataset sizes. {\it Second}, we develop a method, FiSH, short for {\bf F}a{\bf i}r {\bf S}patial {\bf H}ot Spots, for efficiently enumerating sets of hot spots along the quality-fairness trade-off. FiSH works as a layer over any chosen fairness-agnostic spatial hot spot detection method, making it available across diverse scenarios and existing methodologies for those scenarios. {\it Third}, we outline a suite of evaluation measures to assess the quality and fairness of results. {\it Lastly}, we perform an extensive empirical evaluation over real-world datasets which illustrates the effectiveness and efficiency of FiSH in identifying diverse and fair hot spots. 

\section{Related Work}

Given that fairness in spatial hot spots is a novel problem, we consider related work across two streams. We start by considering work on identifying {\it outliers} and {\it spatial hot spots}. These tasks are starkly different in terms of how the results are characterized. Outliers are determined based on neighborhood density, whereas hot spots are determined based on {\it hotness} on a chosen attribute (e.g., diseased, poor etc.). In particular, the notion of a hotness attribute is absent in the formulation for outlier detection making them fundamentally different tasks. Despite being non-identical tasks, there are similarities in the overall spirit of the tasks, which makes outlier identification relevant to the interested reader. We start with a discussion on methods for the tasks of {\it outlier detection} and {\it spatial hot spots}, and then move on to work on fairness in machine learning as applied to tasks allied to ours. 


\subsection{Outlier Identification}

There have been a variety of problem settings that seek to identify objects that are distinct from either their surroundings or the broader dataset. The more popular formulations use the former notion, that of measuring contrast from the surroundings of the data object, i.e.,  making use of neighborhood density. LOF~\cite{breunig2000lof} (and improvements~\cite{kriegel2009loop}) consider identifying individual objects, aka {\it outliers}, which are interesting due to their (relatively sparser) spatial neighborhoods. It is noteworthy that these make object-level decisions informed purely by spatial attributes (without reference to non-spatial attributes like diseased/non-diseased, as required for COVID-19 hot spot determination). SLOM~\cite{chawla2006slom} extends the object-level local neighborhood-based decision making framework to incorporate information from non-spatial attributes. Among outlier detection methods that assess the contrast of individual data objects with the dataset as a whole, the popular paradigm is to build a dataset level statistical model, followed by assessing the conformance of individual objects to the model; those that are less conformant would be regarded as outliers. Such statistical models could be a clustering~\cite{yu2002findout}, dirichlet mixture~\cite{fan2011unsupervised}, or more recently, auto-encoders~\cite{chen2017outlier,lai2019robust}. 

\subsection{Spatial Hot Spots}

Spatial Scan Statistics (SSS), pioneered by Kulldorff~\cite{kulldorff1997spatial}, are methods that identify {\it localized regions} that encompass multiple objects (in contrast to making decisions on individual objects, as in LOF) which {\it collectively} differ from overall dataset on chosen non-spatial hotness attributes (e.g. diseased, poor etc.). The focus is on characterizing regions which may be interpreted as {\it hot spots} due to the divergence of their character from the overall dataset. This makes SSS a markedly different task from outlier identification in specification, input data requirements, internal workings and output format. SSS spatial hot spots are vetted using a statistical likelihood ratio test to ascertain significant divergence from global character. This makes SSS as well as its various variants, as implemented within SaTScan, a statistically principled family of methods to detect spatial hot spots. While Kulldorff's initial proposal is designed to identify circular hot spots, the framework has been generalized to identify arbitrary shapes in several ways; ULS~\cite{patil2004upper} is a notable work along that direction. Other methods such as bump hunting~\cite{friedman1999bump} and LHA~\cite{telang2014detecting} address related problems and leverage data mining methods. Despite an array of diverse research in identifying spatial hot spots, SSS methods have remained extremely popular. Just since 2020, there have been 1000+ papers\footnote{https://scholar.google.com/scholar?as\_ylo=2020\&q=satscan\&hl=en\&as\_sdt=0,5} that make use of SSS and other scan statistics within SaTScan. Our technique, FiSH, can work alongside any method that can provide an ordered output of hot spots, such as SaTScan methodologies. 

\subsection{Fair Unsupervised Learning}

While most attention within the flourishing field of fairness in machine learning~\cite{10.1145/3376898} has focused on supervised learning tasks, there has been some recent interest in fairness for unsupervised learning tasks~\cite{DBLP:conf/cikm/0001JV20}. Among the two streams of fairness explored in ML, viz., individual and group fairness (refer~\cite{binns2020apparent} for a critical comparative analysis), most work on fair unsupervised learning has focused on group fairness. Group fairness targets to ensure that the outcomes of the analytics task (e.g., clusters, top-$k$ results etc.) embody a fair distribution of groups defined on protected attributes such as gender, ethnicity, language, religion, nationality or others. As alluded to earlier, the most common instantiation of group fairness has been through the computational notion of {\it statistical parity}, initially introduced within the context of classification~\cite{10.1145/2090236.2090255}. Group fair unsupervised learning work includes those on fair clustering (e.g.,~\cite{chierichetti2017fair}), retrieval (e.g.,~\cite{zehlike2017fa}) and representation learning (e.g.,~\cite{olfat2019convex}). While there has been no work on fair spatial hot spots yet, there has been some recent work on fairness in outlier detection which we discuss below.

\noindent{\bf Fair Outliers:} There has been some recent work on fair outlier detection. We start by outlining the core differences between outlier detection to illustrate why fairness enhancements targeted at outlier detection would not be applicable for spatial hot spots. First, outlier detection involves object-level decision making, whereas hot spots are determined at the level of object groups. Second, they do not make use of any non-spatial hotness attribute (e.g., diseased, poor etc.) to determine outlierness, whereas a {\it key non-spatial attribute is used to characterize hot spots}. The second difference makes algorithms for outlier detection contrast highly from those for identifying spatial hot spots. Among recent fair outlier detection papers,~\cite{fairod2020} develops a human-in-the-loop method for fair outlier detection, whereas~\cite{deepak2020fair} focuses on automated group fair outlier detection, developing {\it FairLOF}, a technique that extends LOF (discussed above) for fairness. FairLOF adapts LOF to incorporate adjustments based on protected attribute memberships of the object in question and its neighbors, to ensure that protected groups are fairly represented among outliers. It may be noted that the protected attributes are used exclusively to embed fairness, and not to characterize outlierness. There is a third paper, ~\cite{shekhar2020fairod} which makes an intriguing proposition of achieving group fairness (on protected attributes) while being unaware of protected attributes at decision time. To our best knowledge, there has been no prior work on fairness in detecting spatial hot spots or anomalous object groups of other kinds. 

\section{Problem Statement}

Consider a dataset $\mathcal{D} = \{ \ldots, D, \ldots \}$. Each object $D$ is associated with a set of spatial attributes such as $(x,y)$ for a 2D space, or $(lat, long)$ for locations of people. Further, each $D$ is associated with a non-spatial {\it hotness} attribute $v \in \{0,1\}$ such as {\it diseased} (for epidemiology) or {\it poor} (for human development), which is used to determine spatial hot spots. $D$ is also associated with protected attributes (e.g., ethnicity, religion) as we will see momentarily. 


Consider a method for detecting spatial hot spots, such as spatial scan statistics, that is capable of providing a ranked list of top spatial hotspots, as $\mathcal{S} = [ S_1, S_2, \ldots , S_m]$. Each $S_i$ is associated with a spatial region $R_i$ (circular/spherical in the case of the basic SSS) such that the data objects (from $\mathcal{D}$) that fall within $R_i$ have a significantly different hotness profile than the overall dataset. For example, the population within $R_i$ may have a significant high (or low) incidence rate of poverty as compared to the whole population. Noteworthiness of spatial hot spots, analyzed statistically (as in SSS), is directly related to both the size of the population within the hot spot and the extent of divergence on the hotness attribute. $\mathcal{S}$ is the list of spatial hot spots ordered in decreasing statistical noteworthiness; thus $S_i$ is more noteworthy than $S_{i+1}$. When $k$ (typically, $k<<m$) noteworthy spatial hot spots are to be chosen to action upon {\it without consideration to fairness}, the most noteworthy $k$ hot spots, i.e., $\mathcal{S}_{topk} = [S_1, \ldots, S_k]$, would be a natural choice.


\subsection{Fair Spatial Hot Spots} 

The task of fair spatial hot spots detection is to ensure that the $k$ hot spots chosen for policy action, in addition to noteworthiness considerations as above, together encompass a diverse population when profiled along protected attributes such as ethnicity, religion, nationality etc, as motivated earlier. In other words, each demographic group is to be accorded a fair share within the collective population across the chosen hot spots. As mentioned earlier, this notion of {\it statistical parity} has been widely used as the natural measure of fairness in unsupervised learning~\cite{chierichetti2017fair,deepak2020fair,DBLP:conf/nips/BeraCFN19}. When the protected attributes are chosen as those that an individual has no opportunity to actively decide for herself (observe that this is the case with ethnicity, gender as well as religion and nationality to lesser extents), statistical parity aligns particularly well with the philosophy of {\it luck egalitarianism}~\cite{knight2013luck}, as noted in Section~\ref{sec:fairhotspots}. 

We will use $\mathcal{S}_{fairk}$ to denote a set of $k$ hot spots (from $\mathcal{S}$) that are selected in a fairness-conscious way. It is desired that $\mathcal{S}_{fairk}$ fares well on {\it both} the following measures: 

\begin{equation}
    N(\mathcal{S}_{fairk}) = \sum_{S \in \mathcal{S}_{fairk}} rank_{\mathcal{S}}(S)
\end{equation}

\begin{equation}
    F(\mathcal{S}_{fairk}) = \sum_{P \in \mathcal{P}} Div_P(\mathcal{D}, \cup_{S \in \mathcal{S}_{fairk}} Pop(S) )
\end{equation}

The first, $N(.)$, relates with noteworthiness and is simply the sum of the ranks (ranks within $\mathcal{S}$) of the chosen spatial hot spots. Lower values of $N(.)$ are desirable, and $\mathcal{S}_{topk}$ scores best on $N(.)$, due to comprising the top-$k$ (so, $N(\mathcal{S}_{topk}) = \sum_{i=1}^k i = \frac{k \times (k+1)}{2}$). The second, $F(.)$, is a fairness measure, which requires that the population covered across the hot spots within $\mathcal{S}_{fairk}$ be minimally divergent to the overall population, when measured on a pre-specified set of protected attributes $\mathcal{P}$ (e.g., ethnicity, gender); $Div_P(.,.)$ measures divergence on attribute $P \in \mathcal{P}$. The divergence may be computed using a suitable distance measure; we will use Wasserstein distance \cite{vallender1974calculation}. As in the case for $N(.)$, lower values of $F(.)$ are desirable too. Though lower, and not higher, values of $N(.)$ and $F(.)$ indicate deeper noteworthiness and fairness, we refer to these measures as noteworthiness and fairness to avoid introducing new terminology.

\begin{figure*}
  \centering
  \includegraphics[width=0.75\textwidth]{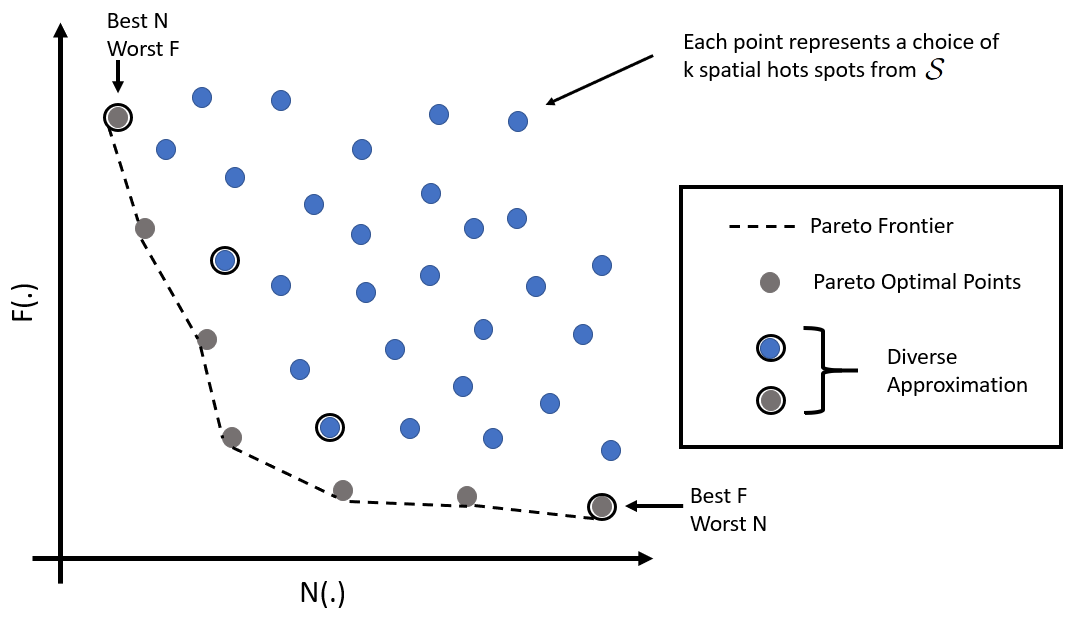}
  \caption{Illustration of the N-F space with $k$-sized subsets of $\mathcal{S}$. The pareto frontier is marked with a dotted line. The circled points indicates a possible solution to the approximate $\tau$-dpe problem ($\tau = 4$). The exact $\tau$-dpe would comprise equally spaced points from the pareto frontier.}
  \label{fig:nfspace}
\end{figure*}

\subsection{Diverse Selection of $\mathcal{S}_{fairk}$ Candidates} 

The noteworthiness and fairness considerations are expected to be in tension (an instance of the fairness-accuracy tension~\cite{menon2018cost}), since {\it fairness is not expected to come for free} (as argued extensively in~\cite{kearns2019ethical}). One can envision a range of possibilities for $\mathcal{S}_{fairk}$, each of which choose a different point in the trade-off between $N(.)$ and $F(.)$. At one end is the $\mathcal{S}_{topk}$ (best $N(.)$, likely worst $F(.)$), and the other end is a maximally fair configuration that may include extremely low-ranked hot spots from $\mathcal{S}$. These would form the pareto frontier\footnote{https://en.wikipedia.org/wiki/Pareto\_efficiency\#Pareto\_frontier} when all the $^{m}C_k$ ($k$ sized) subsets of $\mathcal{S}$ are visualized as points in the 2D noteworthiness-fairness space, as illustrated in Figure~\ref{fig:nfspace}. Each point in the pareto frontier (often called skyline~\cite{borzsony2001skyline}) is said to be {\it pareto efficient} or {\it pareto optimal} since there is no realizable point which is strictly better than it on \underline{both} N and F measures. In other words, $\mathcal{S}_{fairk}$ candidates that are not part of the pareto frontier can be safely excluded from consideration, since there would be a pareto frontier candidate that is strictly better than it on both noteworthiness and fairness. 

Each policy domain may choose a different point in the trade-off offered across candidates in the pareto frontier, after due consideration of several available trade-off points. For example, policing may require a high-degree of fairness, whereas epidemiology interventions may be able to justify policy actions on less diverse populations based on the extent of supporting medical evidence. The pareto frontier may be large (could contain hundreds of candidates, theoretically bounded above only by $\mathcal{O}(^{m}C_k)$) for a human user to fully peruse. Thus, an obvious recourse would be to identify {\it $\tau$ diverse pareto efficient candidates} (henceforth, $\tau$-dpe), where $\tau$ is a pre-specified parameter, so the human user may be able to choose appropriately from a varied set of possibilities. A natural and simple but incredibly inefficient solution would be to (i) enumerate the entire pareto frontier, (ii) trace the sequence of pareto efficient points from the top-left to the bottom-right (i.e., the dotted line), (iii) split the sequence into $\tau - 1$ equally sized segments, and (iv) take the $\tau$ segment end points as the result. 

To summarize, the diverse candidate selection task outlined as $\tau$-dpe {\it requires a diverse set of pareto efficient candidates in the N-F space, each candidate representing a $k$ sized subset of} $\mathcal{S}$. 

\subsection{Approximate $\tau$-dpe} 

It may be observed that it is infeasible to enumerate the $^mC_k$ subsets (e.g., $^{40}C_{10} = 8.5E$+$8$) in the N-F space just due to the profusion of possibilities, making exact $\tau$-dpe identification (as outlined in the four-step process in the previous section) infeasible for practical scenarios. This makes the task of identifying a close approximation of $\tau$-dpe results efficiently a natural alternative for a policy expert to examine the trade-off points and arrive at a reasonable choice of $\mathcal{S}_{fairk}$ to subject to policy action. This brings us to the {\it approximate $\tau$-dpe} task, which is that of efficiently identifying a close approximation of the exact $\tau$-dpe result. The set of circled points in Figure~\ref{fig:nfspace} illustrates a possible solution to the approximate $\tau$-dpe task. All pertinent notations are outlined in Table~\ref{tab:notations} for easy reference. Our method, {\it FiSH}, that addresses the approximate $\tau$-dpe task, is detailed below.

\begin{table}[]
    \centering
    \begin{tabular}{cl}
    \hline
    Notation & What it stands for \\
    \hline
    \hline
        $\mathcal{S}$ & the ordered list of spatial hotspots used as  \\
         & starting point for $\tau$-dpe task\\
        $\mathcal{S}_{topk}$ & the subset of $k$ most noteworthy hotspots from $\mathcal{S}$ \\
        $\mathcal{S}_{fairk}$ & $k$-sized subset of $\mathcal{S}$, a candidate for fair \\
        & selection of hot spots \\
        $N(\mathcal{S}_{fairk})$ & sum of ranks of the spatial hot spots within \\
        & $\mathcal{S}_{fairk}$; lower denotes better noteworthiness \\
        $F(\mathcal{S}_{fairk})$ & deviation of $\mathcal{S}_{fairk}$'s population from  dataset on \\ 
        & protected attributes; lower denotes better fairness \\
        $m$ & cardinality of $\mathcal{S}$ \\
        $k$ & \# hotspots from $\mathcal{S}$ desired in each output candidate \\
        $\tau$ & Number of candidates desired in output \\
        $b$ & beam width parameter used by {\it FiSH} (Sec~\ref{sec:fish}) \\
    \hline     
    \end{tabular}
    \caption{Table of Notations for Easy Reference}
    \label{tab:notations}
\end{table}

\section{FiSH: Fair Spatial Hot Spots}\label{sec:fish}

{\it FiSH} is an efficient heuristic-driven technique addressing the {\it approximate $\tau$-dpe} task outlined above. We first describe a systematic organization of the search space, followed by a heuristic method that traverses the space prioritizing the search using three considerations: {\it pareto efficiency}, {\it diversity} and {\it efficient search}. 

\begin{figure*}
  \centering
  \includegraphics[width=8cm]{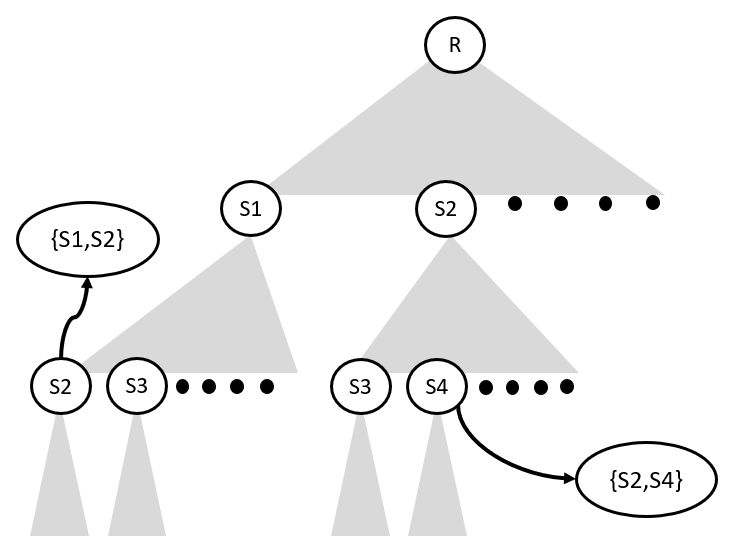}
  \caption{FiSH's Search Tree: Nodes at level $k$ represent $k$ sized subsets of $\mathcal{S}$, and form points in the N-F space (Fig~\ref{fig:nfspace}).}
  \label{fig:tree}
\end{figure*}

\subsection{Search Space Organization} 

Recall that we start with a noteworthiness-ordered list of spatial hot spots $\mathcal{S} = [S_1, \ldots, S_m]$. Our full search space comprises the $^mC_k$ distinct $k$-sized subsets of $\mathcal{S}$. We use the lexical ordering in $\mathcal{S}$ to organize these candidates as leaves of a tree structure, as shown in Figure~\ref{fig:tree}. Each node in the tree is labelled with an element from $\mathcal{S}$, and no node in the FiSH search tree has a child that is lexically prior to itself. Such a hierarchical organization is popular for string matching tasks, where they are called prefix trees~\cite{yazdani2001prefix}. In devising {\it FiSH}, we draw inspiration from using prefix structures for skyline search over databases~\cite{debapriyo2009efficient}. Each internal node at level $l$ (root level $=0$) represents a $l$-sized subset of $\mathcal{S}$ comprising the $l$ nodes indicated in the path from root to itself. The lexical ordering ensures that each subset of $\mathcal{S}$ has a unique position in the tree, one arrived at by following branches corresponding to nodes in the subset according to the lexical ordering. The $^mC_k$ candidates would be the nodes at level $k$. It is infeasible to enumerate them fully, as observed earlier. Thus, {\it FiSH} adopts a heuristic search strategy to traverse the tree selectively to follow paths leading to a good solution (i.e., set of $\tau$ nodes at level $k$) for the approximate $\tau$-dpe task. 

\subsection{FiSH Search Strategy} 

The exact $\tau$-dpe result set is characterized by {\it pareto efficiency} and {\it diversity}, when applied over the $^mC_k$ candidates. The FiSH search strategy uses precisely these criteria as heuristics to traverse the search tree efficiently from the root downward. The core idea behind this search strategy is our conjecture that pareto efficiency and diversity at a given level in the FiSH search tree would be predictive of pareto efficiency and diversity at the next level. We operationalize this heuristic strategy using beam search, a classical memory-optimized search meta-heuristic~\cite{steinbiss1994improvements} that has received much recent attention~\cite{wiseman2016sequence}. 

FiSH starts its search from the root node, expanding to the first-level child nodes, each of which represent singleton-sets denoting the choice of a particular spatial hot spot from $\mathcal{S}$. This forms the candidate set at level 1 of the FiSH tree, $\mathcal{C}_1 = \{ \{S_1\}, \{S_2\}, \ldots \}$. These 1-sized subsets of $\mathcal{S}$ are then arranged in an N-F space as in Fig~\ref{fig:nfspace}. Note that the N-F space of $1$-sized subsets is distinct and different from the N-F space of $k$-sized subsets (Fig.~\ref{fig:nfspace}). The pareto-efficient subset of $\mathcal{C}_1$ is then identified as $P(\mathcal{C}_1)$. The candidates in $P(\mathcal{C}_1)$ are then arranged in a linear sequence tracing the pareto frontier from the top-left to the bottom-right point (similar to the illustration of pareto frontier in Fig~\ref{fig:nfspace}). This linear sequence is split into $b-1$ equally spaced segments, and the $b$ points at the segment end-points are chosen as $D_b(P(\mathcal{C}_1))$, a $b$-sized subset of $\mathcal{C}_1$. The candidate set at the next level of the tree search process, i.e., $\mathcal{C}_2$, is simply the set of all children of nodes in $D_b(P(\mathcal{C}_1))$ (actually, the subsets of $\mathcal{S}$ that they stand for).

\begin{equation}
    \mathcal{C}_2 = \bigcup_{c \in D_b(P(\mathcal{C}_1))} children(c)
\end{equation}

It may be noted that $\mathcal{C}_2$ is a small subset of the set of all $2$-sized subsets of $\mathcal{S}$, since only children of the $b$ nodes selected from the previous level are selected for inclusion in $\mathcal{C}_2$. Next, $\mathcal{C}_2$ is subject to the same processing as $\mathcal{C}_1$ comprising:
\begin{enumerate}
\item identifying pareto efficient candidates $P(\mathcal{C}_2)$, 
\item identifying a diverse $b$ sized subset $D_b(P(\mathcal{C}_2))$, and 
\item following the children pointers, 
\end{enumerate}

to arrive at the candidate set for the next level. This process continues up until $\mathcal{C}_k$ whereby the pareto frontier $P(\mathcal{C}_k)$ is identified followed by the choice of $\tau$ diverse candidates which will eventually form FiSH's result set for the approximate $\tau$-dpe task. This search strategy is illustrated formally in Algorithm~\ref{alg:algo}. The one-to-one correspondence between nodes in the search tree and subsets of $\mathcal{S}$ allows us to use them interchangeably in the pseudocode.

\begin{algorithm}[t]
\SetKwInOut{Input}{input}
\SetKwInOut{Output}{output}
\SetKwInOut{Parameters}{parameters}
\Input{$\mathcal{S}$ organized as a search tree, $k$, $\tau$}
\Parameters{beam width $b$}
$\mathcal{C}_1 = \{\{S_1\}, \{S_2\}, \ldots, \}$ \\
\For{$i\gets1$ \KwTo $k-1$}{
    $P(\mathcal{C}_i) = \text{pareto frontier of }\mathcal{C}_1\text{ in the N-F space}$ \\
    $D_b(P(\mathcal{C}_i)) = \text{equally spaced }b\text{ candidates from pareto frontier }P(\mathcal{C}_i)$ \\
    $\mathcal{C}_{i+1} = \bigcup_{C \in D_b(P(\mathcal{C}_i))} Children(C)$
    }
$P(\mathcal{C}_k) = \text{pareto frontier of }\mathcal{C}_k\text{ in the N-F space}$ \\
$\mathcal{R} = \text{equally spaced }\tau\text{ points from }P(\mathcal{C}_k)$; \\
Return $\mathcal{R}$
\caption{FiSH Search Technique}
\label{alg:algo}
\end{algorithm}

\subsection{Discussion} 

FiSH's search strategy makes use of pareto efficiency and diversity directly towards identifying a small set of nodes to visit at each level of the tree. Restricting the search to only $b$ nodes at {\it each} level before moving to the next enables efficiency. Smaller values of $b$ enable more efficient traversal, but at the cost of risking missing out on nodes that could lead to more worthwhile members of the eventual result set. In other words, a high value of $b$ allows a closer approximation of the $\tau$-dpe result, but at a slower response time. It may be suggested that $b$ be set to $\geq \tau$, since the algorithm can likely afford to visit more options than a human may be able to peruse eventually in the result set. The candidate set size at any point, and thus the memory requirement, is in $\mathcal{O}(bm)$. The computational complexity is in $\mathcal{O}(kb^2m^2)$, and is dominated by the pareto frontier identification (which is in $\mathcal{O}(b^2m^2)$) at each level. While $b$ is a controllable hyperparameter (likely in the range of 5-20), $m$ can be constrained by limiting FiSH to work with the top-$m$ result set (as $\mathcal{S}$) from the upstream spatial hot spot technique. 


\section{Evaluating Approx $\tau$-dpe Results}

Given that (approximate) $\tau$-dpe is a new task we proposed, we now describe novel evaluation metrics to assess the quality of {\it FiSH}'s results. Recall that, given the N-F space comprising all $k$-sized subsets of $\mathcal{S}$, the choice of $\tau$ equally spaced skyline candidates forms the result set for the exact $\tau$-dpe task that we propose in this paper. This result set, which we call {\it Exact}, is computationally infeasible for moderate datasets, but forms our natural baseline for measuring {\it FiSH}'s effectiveness. Approximate $\tau$-dpe results from FiSH may be evaluated either {\it directly based on how well they approximate the expected results of the exact $\tau$-dpe task}, or based on {\it how well they adhere to the spirit of the $\tau$-dpe task} of identifying a diverse group of pareto efficient subsets of $\mathcal{S}$. We now devise evaluation measures along the lines above. In what follows, we use $\mathcal{P}$ to denote the $^mC_k$ $k$-sized subsets of $\mathcal{S}$. 

\subsection{Direct Comparison} 


Let the result of the exact $\tau$-dpe task be $\mathcal{E} = [E_1, \ldots, E_{\tau}]$, and FiSH's result be $\mathcal{F} = [F_1, \ldots, F_{\tau}]$. We would like the average distance between corresponding elements to be as low as possible. 


\begin{equation}
    DC(\mathcal{E}, \mathcal{F}) = \frac{1}{\tau} \sum_{i=1}^{\tau} Eucl(E_i,F_i)
\end{equation}

where $Eucl(.,.)$ is the euclidean distance in the N-F space. Notice that when $\mathcal{E} = \mathcal{F}$, $DC(.,.)$ evaluates to $0.0$. Given that $N(.)$ and $F(.)$ would be in different ranges, we will compute the distance after normalizing both of these to $[0,1]$ across the dataset. As may be obvious, smaller values, i.e., as close to $0.0$ as possible, of $DC(.,.)$ are desirable.

\subsection{Quantifying Pareto-ness: Coverage} 

A diverse and pareto efficient set may be expected to collectively dominate most objects in the $N$-$F$ space. Accordingly, we devise a measure, called {\it coverage}, that measures the fraction of candidates in $\mathcal{P}$ that are pareto dominated by at least one candidate in $\mathcal{F}$.

\begin{equation}
    Cov(\mathcal{F}) = \frac{1}{|\mathcal{P}|} \sum_{P \in \mathcal{P}} \mathbb{I}(\exists F \in \mathcal{F}| F \succ P)
\end{equation}

where $F \succ P$ is true when $F$ pareto dominates $P$. A point pareto dominates another if the latter is no better than the former on both attributes, excluding the case where both are identical in terms of their N-F co-ordinates. A candidate being dominated by another indicates that the latter characterizes an absolutely better trade-off point than the former (on both $N(.)$ and $F(.)$). Thus, we would like the result set to be in a way that most, if not all, candidates are dominated by one or more candidates in the result set. $Cov(.)$ is measured as a fraction of the candidates dominated, hence it is in the range $[0,1]$. Full coverage (i.e., $Cov(.) = 1.0$) may not be attainable given that only $\tau$ candidates can be chosen in the result; instead, if we were to choose the entire skyline, we would get $Cov=1.0$ by design. Thus, the extent to which $Cov(\mathcal{F})$ (FiSH's coverage) approaches $Cov(\mathcal{E})$ (coverage attained by the \textit{exact} result) is a measure of FiSH's quality. Coverage, being modelled using pareto domination, may be seen as modelling {\it pareto-ness} of FiSH's result.

\subsection{Diversity of Results}\label{sec:md}

Given that our formulation of the approximate $\tau$-dpe task hinges on the idea that the candidates should be diverse (so that they may embody a variety of different trade-off points), diversity is a key aspect to measure the adherence of the solution to the spirit of the approximate $\tau$-dpe task. We model diversity as the minimum among pairwise distances between candidates in $\mathcal{F}$. 

\begin{equation}
    MD(\mathcal{F}) = min \{ Eucl(F_i,F_j) | \{ F_i, F_j \} \subseteq \mathcal{F}, F_i \neq F_j \}
\end{equation}

Unlike the average of pairwise distances that allows nearby pairs to be compensated by the existence of far away ones, this is a stricter measure of diversity. On the other hand, this is quite brittle, in the sense just one pair of results being proximal would cause $MD(.)$ to go down significantly; in such cases, the $MD(.)$ would not be that representative of the overall diversity in $\mathcal{F}$. Hence, all the evaluation measures must be seen in cognisance of the others. Coming to desirable values of $MD(.)$, we would like $MD(\mathcal{F})$, which measures the lower bound of distances among elements in $\mathcal{F}$, to be as high as possible, and approach the diversity of $\mathcal{E}$, i.e., $MD(\mathcal{E})$. 

\subsection{Discussion}\label{sec:fishdiscussion}

As obvious from the construction, lower values of $DC$, and higher values on both $Cov$ and $MD$ indicate the quality of FiSH's approach. It is also to be seen that $Cov$ and $MD$ should be judged together, since it is easy to maximize coverage without being diverse and vice versa. $Cov$ and $MD$ requires all $^mC_k$ subsets of $\mathcal{S}$ to be enumerated, whereas $DC$ requires additionally that the exact $\tau$-dpe results be computed. {\it This makes these evaluations feasible only in cases where such enumeration can be done}, i.e., reasonably low values of $m$. In addition to the above quality measures, a key performance metric that FiSH seeks to optimize for, is the {\it response time}. %








\section{Experimental Evaluation}

We now describe our empirical study evaluating FiSH. In this section, we describe the dataset used, the experimental setup, our evaluation measures and our experimental results.

\subsection{Dataset and Experimental Setup}

\subsubsection{Dataset} 

We used the Indian Human Development Survey (IHDS)\footnote{\url{https://ihds.umd.edu/data}} dataset, a large-scale survey of India's population conducted in 2011-12. In particular, we used a random sample of 10000 individuals from the data with distinct locations. The location (lat, long) was determined through querying Google Maps based on the district and other location information available in the data. The binary {\it hotness} attribute was chosen as either (i) {\it (annual) income $<$ 100k}\footnote{100k INR is approximately 1.35k\$; India's per capita income is $\approx 2k \$$}, or (ii) {\it education $<$ 2 yrs}. For each setting, we use {\it caste} and {\it religion} as sensitive attributes and {\it low} income/education as hot spot criterion. In other words, we would like to identify a set of spatial hot spots such that the population across them fare poorly on income (education) but religion and caste groups are fairly represented. These choices of attributes for hotness and fairness are abundantly informed by social realities in contemporary India; for example, caste discrimination remains rampant across India, including in urban settlements\footnote{https://www.economist.com/asia/2020/07/23/even-as-india-urbanises-caste-discrimination-remains-rife}. 

\subsubsection{Experimental Setup}\label{sec:exptsetup}

We used SaTScan Bernoulli model to discover hot spots. We implemented {\it FiSH} as well as the {\it Exact} $\tau$-dpe computation (i.e., enumerate all $^mC_k$ subsets, find pareto efficient frontier, and identify $\tau$ diverse subsets) on Python 3 on an Intel 64 bit i5-8265 at 1.6 GHz with 8 GB RAM. Unless otherwise mentioned, we use the following parameter settings: $m=20$, $k=5$ and $\tau = b = 5$. 


\begin{table}
\begin{center}
\begin{tabular}{p{1.1cm}p{1cm}p{0.8cm}p{0.8cm}p{0.8cm}p{1.3cm}} 
 \hline
 {\bf Setting} & {\bf Method} & {\bf DC} $\downarrow$  & {\bf Cov} $\uparrow$ & {\bf MD} $\uparrow$ & {\bf Time(s)} $\downarrow$ \\
 \hline
 \multirow{2}{*}{{\it Income}} & FiSH & 0.112 & 0.995 & 0.034 & \multicolumn{1}{r}{23.11} \\
                         & Exact & N/A & 0.998 & 0.042 & \multicolumn{1}{r}{6536.54} \\  
 \hline
 \multirow{2}{*}{{\it Education}} & FiSH & 0.045 & 0.987 & 0.041 & \multicolumn{1}{r}{23.87} \\
                         & Exact & N/A & 0.997 & 0.081 & \multicolumn{1}{r}{4413.78} \\  
 \hline
\end{tabular}
\caption{Comparative Results (Task setting: $\tau = 5$, $k=5$, $m=20$ and Parameter Setting: $b=5$ for FiSH); arrows denote whether low or high values are desirable.}
\label{tab:companalysis}
\end{center}
\end{table}

\begin{table}
  \begin{tabular}{p{0.8cm}p{1cm}p{1.5cm}} 
  \hline
  \multicolumn{3}{c}{{\it Education}} \\
  \hline
  $m$ & \multicolumn{1}{r}{{\it FiSH}} & \multicolumn{1}{r}{{\it Exact}} \\
  \hline
  15 & \multicolumn{1}{r}{17.83} & \multicolumn{1}{r}{840.37} \\
  20 & \multicolumn{1}{r}{23.87} & \multicolumn{1}{r}{4413.78} \\
  25 & \multicolumn{1}{r}{39.46} & \multicolumn{1}{r}{33151.91} \\
  30 & \multicolumn{1}{r}{49.28} & \cellcolor[gray]{0.5} \\
  35 & \multicolumn{1}{r}{61.49} & \cellcolor[gray]{0.5} \\
  40 & \multicolumn{1}{r}{71.09} & \cellcolor[gray]{0.5} \\
  \hline
  \end{tabular}
  \caption{Scalability Analysis: Running Time (in seconds) with varying $m$; {\it Exact} did not complete in reasonable time for $m>25$.}
  \label{tab:scalability}
\end{table}

\subsection{Overall Comparison}\label{sec:overallcomp}

\begin{figure*}[!h]
    \centering
    \begin{minipage}{0.3\textwidth}
        \centering
        \includegraphics[width=\textwidth,height = 4cm]{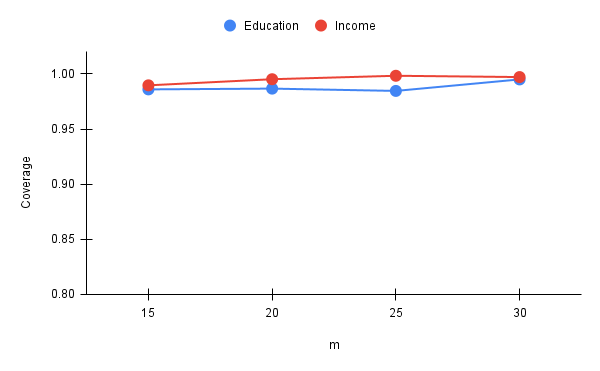} 
        \caption{{\it Cov} vs. m}
        \label{fig:covm}
    \end{minipage}\hfill
    \begin{minipage}{0.3\textwidth}
        \centering
        \includegraphics[width=\textwidth,height = 4cm]{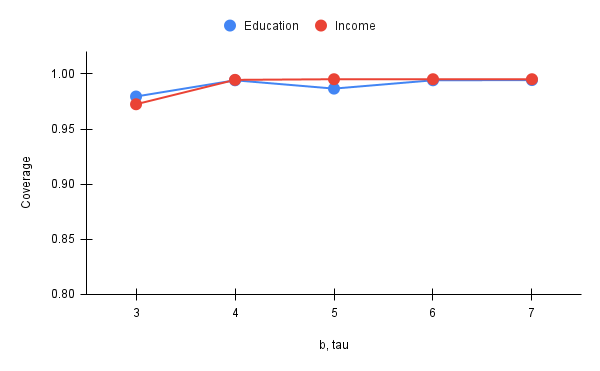} 
        \caption{{\it Cov} vs. b,tau}
        \label{fig:covbt}
    \end{minipage}\hfill
    \begin{minipage}{0.3\textwidth}
        \centering
        \includegraphics[width=\textwidth,height = 4cm]{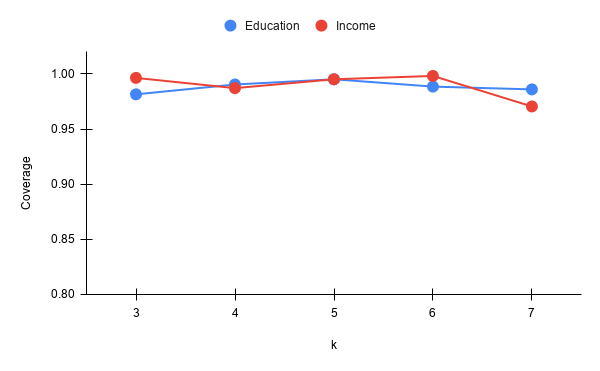} 
        \caption{{\it Cov} vs. k}
        \label{fig:covk}
    \end{minipage}
    \vspace{0.1cm}
    \begin{minipage}{0.3\textwidth}
        \centering
        \includegraphics[width=\textwidth,height = 4cm]{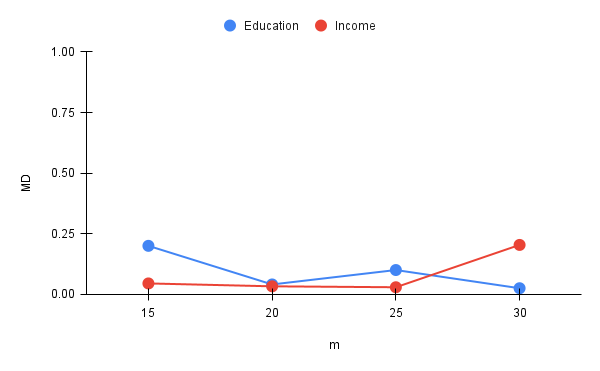} 
        \caption{{\it MD} vs. m}
        \label{fig:mdm}
    \end{minipage}\hfill
    \begin{minipage}{0.3\textwidth}
        \centering
        \includegraphics[width=\textwidth,height = 4cm]{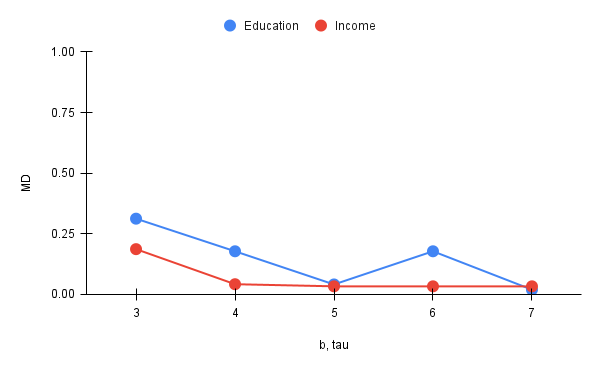} 
        \caption{{\it MD} vs. b,tau}
        \label{fig:mdbt}
    \end{minipage}\hfill
    \begin{minipage}{0.3\textwidth}
        \centering
        \includegraphics[width=\textwidth,height = 4cm]{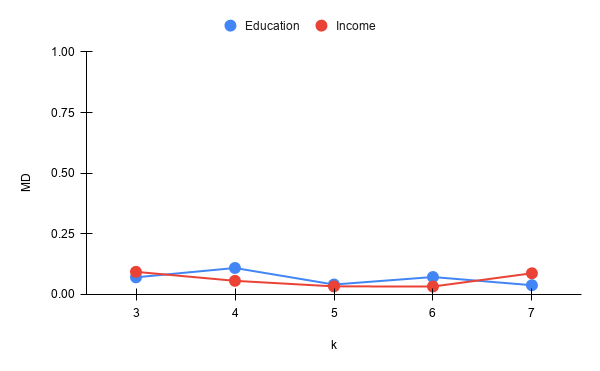} 
        \caption{{\it MD} vs. k}
        \label{fig:mdk}
    \end{minipage}
\end{figure*}

We performed extensive empirical analyses over varying settings. We present representative results and analyses herein. Table~\ref{tab:companalysis} illustrates a representative sample of the overall trends in the comparison between {\it FiSH} and {\it Exact}. The low values of $DC$ indicate that {\it FiSH}'s results are quite close to those of {\it Exact}, which is further illustrated by the trends on the {\it Cov} measure where {\it FiSH} follows {\it Exact} closely. For {\it MD}, we observe a 20\% deterioration in the case of {\it Income}, and a 50\% deterioration in the case of {\it Education}. We looked at the case of {\it Education} and found that the low value of {\it MD} for {\it FiSH} was due to one pair being quite  similar (distance of $0.041$), possibly a chance occurrence that coincided with this setting; the second least distance was more than three times higher, at $0.1349$. On an average, the pairwise distances for {\it FiSH} was only 20\% less than that for {\it Exact}. Across varying parameter settings, a 15-20\% deterioration of {\it MD} was observed for {\it FiSH} vis-a-vis {\it Exact}. For the record, we note that the choice of first $k$ hot spots from $\mathcal{S}$ as the result yielded $DC \approx 0.8$ and {\it Cov} $3$ to $10$ percentage points lower; this confirms that $\tau$-dpe task formulation is significantly different from {\it top-k} not just analytically, but empirically too.

Apart from being able to approximate the {\it Exact} results well, {\it FiSH} is also seen to be able to generate results exceptionally faster, a key point to note given that bringing the $\tau$-dpe task into the realm of computational feasibility was our main motivation in devising {\it FiSH}. In particular, {\it FiSH}'s sub-minute response times compare extremely favourably against those of {\it Exact} which is seen to take more than an hour; we will illustrate later that {\it Exact} scales poorly and rapidly becomes infeasible for usage within most practical real-life scenarios. 

The {\it FiSH} vs. {\it Exact} trends, reported in Table~\ref{tab:companalysis} is representative of results across variations in parameter settings. {\it FiSH} was consistently seen to record 0-10\% deteriorations in {\it Cov}, around 15-25\% deterioration in {\it MD}, and multiple orders of magnitude improvements in response time. The trends on the effectiveness measures as well as the response time underline the effectiveness of the design of the {\it FiSH} method. 

\subsection{Scalability Analysis}

With {\it FiSH} being designed for efficient computation of a reasonable approximation of $\tau$-dpe results, it is critical to ensure that {\it FiSH} scales with larger $m$; recall that $m=|\mathcal{S}|$, the size of the initial list of hotspots chosen to work upon. Table~\ref{tab:scalability} illustrates the {\it FiSH} and {\it Exact} response times with varying $m$. While {\it Exact} failed to complete in reasonable time (we set a timeout to 12 hours) for $m>25$, {\it FiSH} was seen to scale well with $m$, producing results many orders of magnitude faster than {\it Exact}. In particular, it was seen to finish its computation in a few minutes even for $m \approx 100$, which is highly promising in terms of applicability for practical scenarios. Similar trends were obtained with scalability with higher values of $k$ and $\tau$; {\it Exact} quickly becomes infeasible, whereas {\it FiSH}'s response time grows gradually. 

\subsection{Analysis over Varying Settings}

We now analyze the performance of {\it FiSH} in varying settings. This analysis helps us evaluate the sensitivity of {\it FiSH} to specific parameter values; for example, smooth movements along small variations in parameter values will help build confidence in the utility of {\it FiSH} in varying scenarios. With {\it Exact} being unable to complete running within reasonable amounts of time for higher search spaces (e.g., $m>25$, $k=7$, $\tau > 5$ etc.), we restrict our attention to {\it FiSH} trends over {\it Cov} and {\it MD}; this is so since results from {\it Exact} are necessary to compute the {\it DC} measure. Among {\it Cov} and {\it MD}, our expectation is that the brittleness of the {\it MD} measure, as noted in Section~\ref{sec:md}, could lead to more fluctuations in {\it MD} when compared to {\it Cov}, even when {\it FiSH} results change only gradually. We now study the trends with varying parameter settings, changing parameters one at a time, keeping all parameters at their reference settings from Section~\ref{sec:exptsetup}, except the one being varied. 

\subsubsection{Varying $m$} 

We now analyze the effectiveness of {\it FiSH} when operating over a larger set of SaTScan results, i.e., with larger values of $m$ (recall $m=|\mathcal{S}|$). With the number of points in the N-F space being $^mC_k$, increases in $m$ lead rapidly to much denser N-F spaces, and correspondingly larger search spaces. We vary $m$ from $15$ to $30$ in steps of $5$; the {\it Cov} and {\it MD} trends appear in Figure~\ref{fig:covm} and Figure~\ref{fig:mdm} respectively. As expected, {\it Cov} consistently remains at values higher than $0.985$, whereas there is higher volatility in the case of $MD$. The trends indicate that {\it FiSH} is not highly sensitive to $m$ and the quality of its results varies gradually with varying values of $m$. 



\subsubsection{Varying $\tau$}

The number of trade-off points that is provided to the user, or $\tau$, is another important parameter in the $\tau$-dpe task. The beam size in {\it FiSH}, as observed earlier in Section~\ref{sec:fishdiscussion}, is intimately related to $\tau$, and may be expected to be set such that $b \geq \tau$. Higher values of $b$ yield better results at the cost of slower responses; we consistently set $b=\tau$ in our result quality analysis. Higher values of $\tau$ enable choosing more points from the N-F space in the output, and this provides an opportunity to improve on {\it Cov}. However, choosing more points obviously would lead to deterioration in the {\it MD} measure that measures the minimum of pairwise distances. We vary $\tau$ (and thus $b$) from $3$ to $7$, and plot the {\it Cov} and {\it MD} trends in Figures~\ref{fig:covbt} and~\ref{fig:mdbt} respectively, which show gentle and consistent variations. As expected, {\it Cov} is seen to improve and saturate close to the upper bound of $1.0$. {\it MD} on the other hand, is seen to deteriorate but stabilizes soon; the patterns are consistent except for the case of $\tau=5$ for {\it Education}, likely a chance occurrence as analyzed in Section~\ref{sec:overallcomp}.  

\subsubsection{Varying $k$}

The third parameter of importance for the $\tau$-dpe task is $k$, which denotes the number of hotspots to be chosen within each trade-off point in the result. Increasing values of $k$ (up to $m/2$) lead to larger number of points in the N-F space. With the number of trade-off points to be output pegged at $\tau$, achieving the same coverage would become harder with increasing $k$. This is in contrast with $MD$ where there is no expectation of a consistent deterioration or improvement. From the {\it Cov} and {\it MD} plots in Figures~\ref{fig:covk} and~\ref{fig:mdk}, the {\it Cov} is quite stable with a deterioration kicking in at $k=7$ (even there, $Cov$ remains at $0.90+$), whereas {\it MD} remains consistent. 

\subsubsection{Setting $b$} 

The beam width, $b$ in {\it FiSH}, offers a mechanism to trade-off effectiveness for efficiency. We experimented with varying values of $b$ and found that the gains on effectiveness measures (i.e., {\it DC}, {\it Cov} and {\it MD}) taper off beyond $b > 2 \times \tau$. The response times were seen to increase with $b$; there are two ways in which $b$ affects the complexity, one is by providing more candidates at each level (which increases linearly with $b$), and another by increasing the cost of pareto frontier identification (which is in $\mathcal{O}(b^2)$). From the trends which indicated a linear trend between response time and $b$, it may be reasonably suspected that the former factor dominates. 

\begin{figure}
  \centering
  \includegraphics[width=\columnwidth]{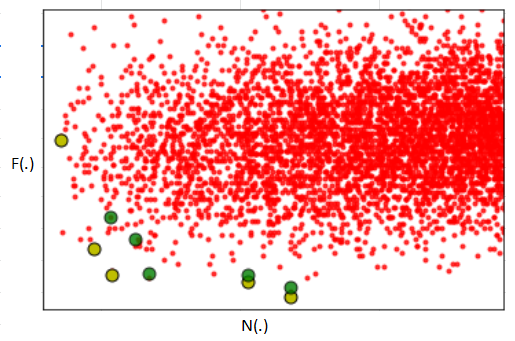}
  \caption{Example Results; kindly view in color. {it FiSH} results in green and {\it Exact} results in mustard yellow.}
  \label{fig:qual}
  \vspace{-0.2in}
\end{figure}

\subsection{\bf Example Results in the N-F Space} 

Having analyzed {\it FiSH} quantitatively, we now consider a qualitative evaluation of {\it FiSH} vis-a-vis {\it Exact}. Fig~\ref{fig:qual} illustrates the N-F space for our reference setting (Section~\ref{sec:exptsetup}) for {\it Income}, with results from {\it FiSH} (green points) juxtaposed against {\it Exact} results (mustard yellow) and other points in red. This result is representative of FiSH's strengths and weaknesses. While three of five FiSH results are seen to be on the pareto frontier, the others are only slightly inward. As in the case of any heuristic-driven method, FiSH may miss some good results; here, FiSH's sampling misses out on the top-left region of the pareto frontier, which explains the slight deterioration in {\it Cov} for {\it FiSH} when compared with {\it Exact}.

\section{Conclusions and Future Work}


In this paper, for the first time to our best knowledge, we considered the task of fair detection of spatial hot spots. In this web era where spatially-anchored digital data is collected extensively, spatial hot spot detection is used extensively to inform substantive policy interventions across a variety of domains, making fairness an important consideration within them. We characterized fairness using the popular notion of statistical parity when computed collectively over $k$ chosen hot spots, and outlined the task of identifying a diverse set of solution candidates along the fairness-noteworthiness pareto frontier. Observing the computational infeasibility of identifying exact solutions, we developed a method, {\it FiSH}, that performs a highly efficient heuristic-driven search to identify good quality approximate solutions for the task. We then formulated a suite of evaluation metrics for the novel task of fair hot spots. We perform an extensive empirical evaluation over a real-world dataset from the human development domain where fairness may be considered indispensable, and illustrated that FiSH delivers high-quality results, and offers good scalability, consistently returning results orders of magnitude faster than what is required to compute exact results. This illustrates the effectiveness of FiSH in achieving fairness in detection of spatial hot spots, and that it offers fast response times, making it appropriate for real-world scenarios. 

\subsection{Future Work} 

While we have considered enhancing fairness by working upon a ranked list of spatial hot spots, FiSH extends easily to work over techniques that are capable of providing scores (in addition to ranks, which is basically an ordering over the scores) for each hot spot as well; we are considering evaluating FiSH's effectiveness in working over such scored lists. Our formulation of diverse candidates assumes that the user may be interested equally in all parts of the noteworthiness-fairness trade-off space. However, in several cases, users may have a preference to exclude some parts of the space. For example, the maximum relaxation of noteworthiness may be bounded above in some scenarios. We are considering how user's trade-off preferences can be factored into the FiSH search process to deliver diverse results within the sub-spectrum of interest.

\bibliographystyle{ACM-Reference-Format}
\bibliography{refs}

\end{document}